\documentclass[conference]{IEEEtran}

\IEEEoverridecommandlockouts

\ifCLASSINFOpdf
\else
\fi

\usepackage{amsmath,amssymb}
\usepackage{graphicx}  
\usepackage{citesort}

\newcommand{\pumpf}             {\omega_p}
\newcommand{\htf}               {{G}}
\newcommand{\htfe}              {\mathbf{\hat{G}}}
\newcommand{\htfen}             {{\hat{G}}}

\newcommand{\forcing}           {f_0(t)}
\newcommand{\chirp}             {u(t)}
\newcommand{\pumping}           {f(t)}

\newcommand{\req}[1]          {(\ref{#1})}
\newcommand{\refig}[1]        {Fig.~\ref{#1}}
\newcommand{\resec}[1]        {Section~\ref{#1}}

\begin{document}
\title{Identification of a Hybrid Spring Mass Damper via Harmonic Transfer Functions as a Step Towards Data-Driven Models for Legged Locomotion}

\author{\IEEEauthorblockN{\.{I}smail Uyan{\i}k\IEEEauthorrefmark{1},
M. Mert Ankaral{\i}\IEEEauthorrefmark{2},
Noah J. Cowan\IEEEauthorrefmark{2},
\"{O}mer Morg\"{u}l\IEEEauthorrefmark{1} and
Ulu\c{c} Saranl{\i}\IEEEauthorrefmark{3}} \\
\IEEEauthorblockA{\IEEEauthorrefmark{1}Dept. of Electrical and Electronics Engineering, Bilkent University, 06800 Ankara, Turkey\\
Email: uyanik@ee.bilkent.edu.tr, morgul@ee.bilkent.edu.tr}
\IEEEauthorblockA{\IEEEauthorrefmark{2}Dept. of Mechanical Engineering, Johns Hopkins University, Baltimore, MD 21218, USA\\
Email: mertankarali@jhu.edu, ncowan@jhu.edu}
\IEEEauthorblockA{\IEEEauthorrefmark{3}Dept.\ of Computer Engineering, Middle East Technical University, 06800 Ankara, Turkey\\
Email: saranli@ceng.metu.edu.tr}}

%\author{\IEEEauthorblockN{Michael Shell}
%\IEEEauthorblockA{School of Electrical and\\Computer Engineering\\
%Georgia Institute of Technology\\
%Atlanta, Georgia 30332--0250\\
%Email: http://www.michaelshell.org/contact.html}
%\and
%\IEEEauthorblockN{Homer Simpson}
%\IEEEauthorblockA{Twentieth Century Fox\\
%\and
%\IEEEauthorblockN{James Kirk\\ and Montgomery Scott}
%\IEEEauthorblockA{Starfleet Academy\\
%San Francisco, California 96678-2391\\

\maketitle

\begin{abstract}

    There are limitations on the extent to which manually constructed
  mathematical models can capture relevant aspects of legged
  locomotion. Even simple models for basic behaviours such as running
  involve non-integrable dynamics, requiring the use of possibly
  inaccurate approximations in the design of model-based
  controllers. In this study, we show how data-driven frequency domain
  system identification methods can be used to obtain input--output
  characteristics for a class of dynamical systems around their limit
  cycles, with hybrid structural properties similar to those observed
  in legged locomotion systems. Under certain assumptions, we can
  approximate hybrid dynamics of such systems around their limit cycle
  as a piecewise smooth linear time periodic system (LTP), further
  approximated as a time-periodic, piecewise LTI system to reduce
  parametric degrees of freedom in the identification process. In this
  paper, we use a simple one-dimensional hybrid model in which a
  limit-cycle is induced through the actions of a linear actuator to
  illustrate the details of our method. We first derive theoretical
  harmonic transfer functions of our example model. We then excite the
  model with small chirp signals to introduce perturbations around its
  limit-cycle and present systematic identification results to
  estimate the harmonic transfer functions for this model. Comparison
  between the data-driven HTF model and its theoretical prediction
  illustrates the potential effectiveness of such empirical
  identification methods in legged locomotion.

\end{abstract}


% \fi
\IEEEpeerreviewmaketitle

%%%%%%%%%%%%%%%%%%%%%%%%%%%%%%%%%%%%%%%%%%%%%%%%%%%%%%%%%%%%%%%%%%%%%%%%%%%%%%%%
\section{Introduction}
\label{sec:introduction}

Legged locomotion emerges from a staggering diversity of animal and
robot morphologies and gaits, and modeling locomotor dynamics remains
a grand challenge in both biology and robotics
\cite{holmesdynamics2006,fulltemplates1999}.  Running behaviors, in
particular, are commonly represented by relatively simple spring--mass
models such as the Spring-Loaded Inverted Pendulum (SLIP) model
\cite{SchwindPhD98,fullmechanics1991}. A common feature of such
models, however, is that their hybrid system dynamics involve
intermittent foot contact with the ground, alternating between flight
and stance phases during locomotion. Despite the presence of seemingly
simple models for basic behaviors such as running and walking, the
hybrid dynamics associated with these behaviors can be rather complex,
with non-integrable parts such as the stance phase
\cite{Holmes90,SchwindPhD98}. Given the utility of having accurate
models and associated analytic solutions in constructing high
performance controllers for nonlinear systems, substantial effort has
been devoted to the construction of approximate solutions to such
non-integrable hybrid models
\cite{schwind.jnls00,geyer.jtb05,ndpaper,ankaralistride2010}.

When accurate analytical solutions to the dynamics of a legged
platform are available \cite{ndpaper}, their structure can be
exploited to yield effective solutions for system identification and
adaptive control \cite{uyanik_saranli_morgul.icra2011}. Despite our
previous studies showing how accurate such models may be, there will
always be unmodeled components in the physical system, resulting in
discrepancies between the model and experiments
\cite{uyanik.tro}. Attempts to manually incorporate these effects into
the model is daunting at best, and often impossible. Consequently, we
propose an alternative method in this study, namely using data-driven
system identification methods to derive an input--output transfer
function for such hybrid legged locomotion behaviors, thereby
eliminating the need to manually construct an explicit mathematical
model for the system.

Our main goal in this study is to provide a system identification
framework applicable to a useful (although not comprehensive) class of
legged locomotion models \cite{ndpaper}, and possibly more complex
robotic systems \cite{saranlirhex:2001}. Our approach is based on
considering legged locomotion as a hybrid nonlinear dynamical system
with a stable periodic orbit (limit-cycle), corresponding to the
locomotor behavior of interest. We introduce a formulation that
addresses the input--output system identification problem in the
frequency domain for a sub-class of hybrid legged locomotion
models. More specifically, following certain assumptions on the hybrid
dynamics of legged systems, we approximate their hybrid dynamics
around the limit-cycle as a linear time-periodic system
(LTP). However, this first LTP approximation is infinite dimensional,
making parametric identification challenging. We hence further
approximate the dynamics as a finite dimensional \textit{piecewise
  LTI} system (maintaining its LTP nature), thereby limiting the
parametric degrees of freedom while enabling a practical
identification framework.

Existing studies on system identification of LTP systems focus on
modeling these systems as multi-input single-output LTI systems. This
approach is based on the concept of harmonic transfer functions
\cite{wereley.phdthesis}, which are infinite-dimensional operators
that are analogous to frequency response functions for LTI systems. An
identification strategy for such systems was developed in
\cite{afreen.msthesis} using power spectral density and cross spectral
density functions. A similar method was used in \cite{hwang.phdthesis}
considering the effects of noise in both input and output
measurements. Different than these studies, local polynomial methods
and lifting approaches were also used for the identification of
harmonic transfer functions for multi-input single-output models of
LTP systems
\cite{allen.nonlinear}. % This method was also extended to identify
%transfer functions of a wind turbine from simulation data in
%\cite{allen.wind}.

Our contributions in this paper focus on representing the dynamics of
legged locomotion as a linear time periodic system, thereby enabling
the use of the system identification method proposed in
\cite{afreen.msthesis} for such systems. We achieve this by using a
new phase definition in identifying the harmonic transfer functions,
illustrated in the context of a simplified model designed to mirror
structural properties of legged locomotion models. When the problem is
approached as a grey-box model with finite parameters (piecewise LTI),
it suffices to non-parametrically estimate a finite number of
harmonics, to which we later fit parametric models.
%
%\njccom{I don't
%  know what the next sentence means.} \uscom{I think we can take it out.} In doing so, we place particular
%emphasis on identifying piecewise LTI systems using LTP system
%identification approach considering the effect of hybrid system
%dynamics.

Ankarali and Cowan \cite{ankaralisystem2014} developed a similar
system identification method for hybrid systems with periodic orbits
using ``discrete time'' harmonic transfer functions.  However, the
framework and assumptions in this paper are distinctly different from
their approach. Specifically, they use mappings between different
cross sections to construct a discrete-time LTP system, and use
discrete time HTFs for identification. Also, the current paper focuses
on harmonic balance, which also distinguishes this paper from
\cite{ankaralisystem2014}.

\section{Representation of Legged Locomotion as a Hybrid Dynamical System}
\label{sec:hybrid_system}

Our goal in this study is to provide a system identification framework
for a class of models related to legged locomotion using harmonic
transfer functions. For the present paper, we limit ourselves to
``clock-driven'' locomotion models as described in \resec{sec:smooth},
representative of controllers used by a wide variety of existing
robots \cite{saranlirhex:2001,gallowayx-rehx:2010,saranlimodular2011},
with open-loop central pattern generators (CPG) coordinating control
actions to achieve time periodic behaviour.  This will allow us to
directly use time periodicity in our LTP analysis, while eliminating a
variety of complications associated with estimating the phase
\cite{revzenestimating2008}.

\subsection{Smooth Clock-driven Oscillators}
	\label{sec:smooth}
	
In general, the dynamics of smooth, clock-driven oscillators with external inputs can be written as
\begin{align}
	\label{eq:smooth}
	\begin{aligned}
		\dot{q} &= f(q,\phi,u), \\
		\dot{\phi} &= 1 \\
		f &: \mathbb{R}^n \times S^1 \times \mathbb{R}^{p} \mapsto
		\mathbb{R}^n \\
		(q , \ \phi) &\in \ \mathbb{R}^n \times S^1 ,
		\ u \in \mathbb{R}^q
	\end{aligned}
\end{align}
\noindent where $(q , \phi)$ and $\mathbb{R}^n \times S^1$ denote the
state vector and the state space of the oscillator, respectively. The
circle component $S^1 = \mathrm{mod}(\mathbb{R}^+,T)$ enforces the
periodicity of the dynamics, while the external input $u(t)$
represents small external perturbations which we will use for system
identification.

In this paper, we focus on oscillators of the form (\ref{eq:smooth})
with asymptotically stable, isolated periodic orbits (limit-cycle)
$\bar{q}(t) = \bar{q}(t-T)$ when $u(t)=0$. For such systems, if we let
$q(t) = \bar{q}(t) + x(t)$ and linearize the dynamics in
(\ref{eq:smooth}) around the limit-cycle $\bar{q}(t)$, and $u(t) = 0$
we get
\begin{align}
  \label{eq:ltp}
  \begin{aligned}
    \dot{x}(t) = A(t) x(t) + B(t) u(t) \\
    y(t) = C(t) x(t) + D(t) u(t)
  \end{aligned}
\end{align}
\noindent where
\begin{eqnarray}
    A(t) &=& \left[ \frac{\partial f}{\partial q} \right]_{ \begin{smallmatrix} q(t) &=& \bar{q}(t)\\ u(t) &=& 0 \end{smallmatrix} }, \\
	B(t) &=& \left[ \frac{\partial f}{\partial u} \right]_{\begin{smallmatrix} q(t) &=& \bar{q}(t)\\ u(t) &=& 0 \end{smallmatrix}}.
\end{eqnarray}
This corresponds to a Linear Time Periodic (LTP) system, with all
system matrices sharing a common period, $T$.

\subsection{Modeling Framework for Hybrid Systems}
\label{sec:modeling_hybrid_systems}

\begin{figure}[b!]
\begin{center}
\includegraphics[width=0.95\columnwidth]{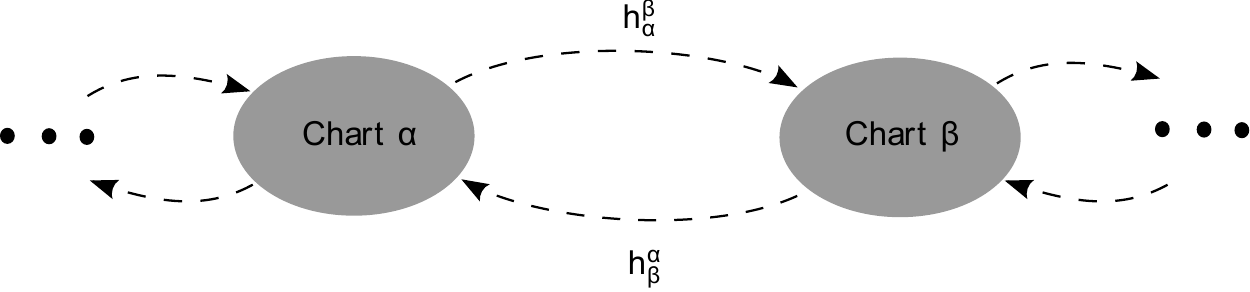}
\caption{A simple state transition graph for a hybrid dynamical system.}
\label{fig:statetransition}
\end{center}
\end{figure}

Legged systems are often modeled using hybrid dynamics due to
intermittent foot contact with the ground, which cannot be represented
with a single, smooth dynamical flow. In the broadest sense, a hybrid
dynamical system is a set of smooth flows together with discrete
transitions (and associated transformations) between these flows
triggered by intersections of system trajectories with sub-manifolds
of the continuous state space \cite{guckenheimerplanar1995}. These
flows are called {\em charts}, indexed with unique labels $\mathcal{I}
:= \lbrace 0 , \cdots , d \rbrace$ each with possibly different
equations of motion. Along its trajectories, a hybrid system
transitions from one chart to another, with each transition defined by
the zero crossing of a {\em threshold function}. For each source chart
$\alpha \in \mathcal{I}$ and destination chart $\beta \in
\mathcal{I}$, the threshold function $h_{\alpha}^{\beta}$ defines the
transition from chart $\alpha$ to chart $\beta$. An example transition
graph for a hybrid dynamical system is illustrated in
\refig{fig:statetransition}.

Since we are interested in the local behaviour around the limit-cycle,
we assume that there is only one transition function associated with
each chart.\footnote{This approach does not apply to gaits such as
  pronking that nominally involve multiple legs making contact with
  the ground at the same time when on the limit cycle, because small
  deviations from the limit cycle can lead to different touch-down
  order between legs, violating our assumption.}  We further assume
that system trajectories are continuous at transitions, meaning that
system states do not experience discrete changes coincident with chart
transitions. As a final note, we assume that the hybrid dynamical
system we consider has an isolated periodic orbit ensuring that chart
transitions within the limit cycle are also periodic and consistent.

It is important to note that these assumptions
are generally satisfied by models of common locomotory behaviors such as running and walking
\cite{ndpaper,wu3_D2013} as well as a wide range of legged robots for
which leg masses are negligible compared to the dynamics of a larger
body \cite{saranlirhex:2001,gallowayx-rehx:2010}. Consequently,
the system identification methods we introduce will remain applicable
to systems other than the simplified example we will present in this
paper.

\subsection{Modeling Legged Locomotion as a Linear Time Periodic System}
\label{ltp_system}

For clarity, we limit our focus in this
section to an example hybrid dynamical system with only two charts,
$\mathcal{I} = \lbrace 0 , 1 \rbrace$, designed to capture stance and flight phases of simple spring-mass models of locomotion. Based on a clock driven assumption, for each $i \in \mathcal{I}$ the continuous
dynamics can be represented with
\begin{align}
  \label{eq:hybrid}
    \begin{aligned}
    \dot{\phi} &= 1 \\
    \dot{q}_i &= f_i(q,\phi,u) \ , \
       \\
      q_i &\in \mathbb{R}^n
    \end{aligned}
\end{align}
and let the associated threshold function be $h_i^{\mathrm{mod}(i+1,2)}(q)$.
The transition map associated with each hybrid event is simply the identity map,
$q_{i} \mapsto q_{i}$, due to the continuity assumption.
Our linearization of these hybrid dynamics towards an
LTP approximation assumes that these transition times, $\hat{t}$, zero crossings of $h_0^1(q)$ and $h_1^0(q)$,
maintain their periodicity and offsets within the period in close proximity of the limit-cycle, resulting in the following form of the nonlinear dynamics
\begin{align}
  \dot{\phi} &= 1 \\
  \dot{q} &\approx \left\lbrace
    \begin{aligned}
      &f_0(q,\phi,u) \ , \
      \mathrm{if} \ \mathrm{mod}(t,T) \in [0,\hat{t})  \\
      &f_1(q,\phi,u) \ , \
      \mathrm{if} \ \mathrm{mod}(t,T) \in [\hat{t},T)
    \end{aligned} \right.\;.
\end{align}
Assuming that the system given above has a limit cycle $\bar{q}(t)$ with a
period $T$, linearization around $\bar{q}(t)$ yields the piecewise smooth LTP system
\begin{align}
  \notag
  \begin{aligned}
    \dot{x}(t) &= \left\lbrace
    \begin{aligned}
      &A_0(t)
      x(t) +B_0(t) u(t)
      , \ \mathrm{if} \  \mathrm{mod}(t,T) \in [0,\hat{t})
    \\
       &A_1(t)
      x(t) +B_1(t) u(t)
      ,  \ \mathrm{if} \ \mathrm{mod}(t,T) \in [\hat{t},T)
    \end{aligned}
    \right.
  \end{aligned}
\end{align}
where
\begin{align}
	\notag
	\begin{aligned}
		A_0(t) := \left[ \frac{\partial f_0}{\partial q} \right]_{ \begin{smallmatrix} q(t) &=& \bar{q}(t)\\ u(t) &=& 0 \end{smallmatrix}  } , B_0(t):=\left[ \frac{\partial f_0}{\partial u} \right]_{ \begin{smallmatrix} q(t) &=& \bar{q}(t)\\ u(t) &=& 0 \end{smallmatrix} } ,\\
		A_1(t):=\left[ \frac{\partial f_1}{\partial q} \right]_{\begin{smallmatrix} q(t) &=& \bar{q}(t)\\ u(t) &=& 0 \end{smallmatrix} } , B_1(t):=\left[ \frac{\partial  f_1}{\partial u} \right]_{\begin{smallmatrix} q(t) &=& \bar{q}(t)\\ u(t) &=& 0 \end{smallmatrix} }.
	\end{aligned}
\end{align}

It is natural
to assume that direct measurement of all $x(t)$ may not be available
or we may only measure a subset of $x(t)$. Consequently, we also define a
time-periodic output equation as in the form \req{eq:outputmeasurement}.

Since system matrices $A_i(t)$, $B_i(t)$, $C_i(t)$ and $D_i(t)$ with
$i \in \lbrace 0 , 1 \rbrace$ are time parametrized functions, the
system has infinite parametric degrees of freedom, making parametric
system identification challenging even when Harmonic Transfer
Functions are used. At this point, we hypothesize that for hybrid
systems, the variability within a chart is small compared to the
change between charts and we approximate the LTP dynamics using a
piecewise LTI approximation that preserves the LTP structure of the
system. The LTP equations of motion then take the form
\begin{align}
  \label{eq:stateequation}
  \begin{aligned}
    \dot{x}(t) &\approx \left\lbrace
    \begin{aligned}
      &A_0
      x(t) +B_0 u(t)
      , \ \mathrm{if} \  \mathrm{mod}(t,T) \in [0,\hat{t})
    \\
       &A_1
      x(t) +B_1 u(t)
      ,  \ \mathrm{if} \ \mathrm{mod}(t,T) \in [\hat{t},T)
    \end{aligned}
    \right.
  \end{aligned}
  \\
  \label{eq:outputmeasurement}
  \begin{aligned}
    y(t) &\approx \left\lbrace
    \begin{aligned}
      &C_0
      x(t) +D_0 u(t)
      , \ \mathrm{if} \  \mathrm{mod}(t,T) \in [0,\hat{t})
    \\
       &C_1
      x(t) + D_1 u(t)
      ,  \ \mathrm{if} \ \mathrm{mod}(t,T) \in [\hat{t},T)
    \end{aligned}
    \right.
  \end{aligned}
\end{align}
The formulation above constitutes the basis of our framework for
analyzing and identifying clock-driven legged locomotion models.

\section{Harmonic Transfer Functions}
\label{sec:htf}

\subsection{Preliminaries and Background}

System identification studies on (stable) LTI systems rely on the fact
that if the input is sinusoidal, then, at steady-state, the output
will also be a sinusoidal signal (at the same frequency but with a
possibly different magnitude and phase). This one-to-one mapping
between input and output signals allows us to characterize the
dynamics in terms of a frequency response function (FRF) also know as
a Bode plot. Unfortunately, this approach does not readily transfer to
LTP systems, which produce output spectra that include multiple
(possibly infinite) harmonics of the input stimuli, each with possibly
different magnitude and phase at steady state.

One ad hoc way to mitigate this is to enforce a one-to-one mapping by
neglecting higher harmonics \cite{leonhard}. However, this assumption
may result in substantial inaccuracies particularly when the influence
of higher harmonics on the response is expected to be
significant. Motivated by this problem, Wereley \cite{wereley.1990}
proposed a linear one-to-one mapping between the coefficients of an
exponentially modulated periodic (EMP) signal at the input of LTP
systems to the coefficients of an EMP signal at their output. This
linear operator that maps the input harmonics to the output harmonics
of an LTP system is called a Harmonic Transfer Function (HTF)
\cite{wereley.phdthesis}.

In the following section, we review the derivation of harmonic transfer
functions as presented in \cite{wereley.phdthesis} and
\cite{mollerstedt.phdthesis}, using the principle of harmonic balance
starting from the state space representation of \req{eq:ltp}.

\subsection{Theoretical Derivation of Harmonic Transfer Functions}
\label{sec:htf_theoretical}

Recall that the system matrices in \req{eq:ltp} are all
$T$-periodic. Consequently, they can be represented by an infinite
Fourier series with pumping frequency $\pumpf = 2 \pi / T$. For the
system matrix $A(t)$, we have
\begin{equation}
A(t) = \sum \limits_{n=-\infty}^\infty A_n e^{j \pumpf n t}\;.
\end{equation}
The matrices  $B(t)$, $C(t)$ and $D(t)$ can be similarly decomposed. In addition, we can also expand the state and output vectors since they are both EMP signals. Substituting these expansions into \req{eq:ltp} and applying the principle of harmonic balance as explained in \cite{wereley.phdthesis}, we obtain the harmonic state space representation as
\begin{equation}
\label{eq:harmonicssm}
\begin{aligned}
s \mathcal{X} &= (\mathcal{A} - \mathcal{N}) \mathcal{X} + \mathcal{B} \mathcal{U}        \\
\mathcal{Y} &= \mathcal{C} \mathcal{X} + \mathcal{D} \mathcal{U},
\end{aligned}
\end{equation}
where the doubly infinite vectors representing the harmonics of the
state, control, and output signals are
\begin{equation}
 \begin{aligned}
\mathcal{X}^T &:= [ \cdots, x_{-2}^T, x_{-1}^T, x_{0}^T, x_{1}^T, x_{2}^T, \cdots ], \\
\mathcal{U}^T &:= [ \cdots, u_{-2}^T, u_{-1}^T, u_{0}^T, u_{1}^T, u_{2}^T, \cdots ], \\
\mathcal{Y}^T &:= [ \cdots, y_{-2}^T, y_{-1}^T, y_{0}^T, y_{1}^T,
y_{2}^T, \cdots ],
 \end{aligned}
\end{equation}
and the doubly infinite input modulation matrix is
\begin{equation}
   \mathcal{N} := \mathrm{blockdiag}\{j n \pumpf I \}, \indent \forall n \in Z,
\end{equation}
which modulates the input
frequency to different harmonic frequencies. Details on the derivations can be found in \cite{wereley.phdthesis}.

The $T$-periodic dynamics matrix, $A(t)$, is expressed in terms of its
complex Fourier coefficients, $\{A_n | n \in \mathbb{Z}\}$, as a
doubly infinite block Toeplitz matrix,
\begin{equation}
\mathcal A =
\begin{bmatrix}
\ddots & \vdots & \vdots & \vdots & \vdots & \vdots &         \\
\cdots & A_0    & A_{-1} & A_{-2} & A_{-3} & A_{-4} & \cdots  \\
\cdots & A_1    & A_0    & A_{-1} & A_{-2} & A_{-3} & \cdots  \\
\cdots & A_2    & A_1    & A_0    & A_{-1} & A_{-2} & \cdots  \\
\cdots & A_3    & A_2    & A_1    & A_0    & A_{-1} & \cdots  \\
\cdots & A_4    & A_3    & A_2    & A_1    & A_0    & \cdots  \\
       & \vdots & \vdots & \vdots & \vdots & \vdots & \ddots
\end{bmatrix},
\end{equation}
with a similar definition for $B(t)$ in terms of its Fourier coefficients
represented by $\{B_n | n \in \mathbb{Z}\}$, $C(t)$ in terms of $\{C_n | n
\in \mathbb{Z}\}$, and $D(t)$ in terms of $\{D_n | n \in \mathbb{Z}\}$.

This collection of doubly infinite matrices is called the harmonic
state space model ($\mathcal{HSS}$) of the system given
in~\req{eq:ltp}. However, it will also be useful to determine an
explicit input--output functional relationship between the Fourier
coefficients of the harmonics of the input, $\{u_n | n \in
\mathbb{Z}\}$, and those of the output, $\{y_n | n \in
\mathbb{Z}\}$. This relationship is represented by the harmonic
transfer function, $\htf(s)$, which is also an infinite dimensional
matrix of Fourier coefficients, satisfying
\begin{equation}
\mathcal{Y} = \htf \mathcal{U}\;.
\end{equation}
Based on \req{eq:harmonicssm}, $\htf$ can be computed as
\begin{equation}
\htf = \mathcal{C} [s I - (\mathcal{A - N})]^{-1} \mathcal{B} + \mathcal{D}\;,
\end{equation}
as long as the inverse within this equation exists.

There are, however, two problems associated with the harmonic transfer
function as stated above. First, it is not clear whether the inverse
of the doubly infinite matrix in the definition of the harmonic
transfer function will always exist. This problem will be dealt with
by an application of the Floquet Theorem. Second, the harmonic
transfer function is a doubly infinite matrix operator, which cannot
practically be implemented on a computer. This second problem will be
mitigated by truncating the HTF in order to implement analysis on a
computer.

Note that the theoretical definition of harmonic transfer functions in
\cite{wereley.phdthesis}, reviewed in this section, requires the state
space representation of the system to be available. Our goal is to
estimate this theoretically computed transfer function $\htf (s)$ by
using input-output data in the frequency domain without necessitating
knowledge of internal system dynamics.

\subsection{Estimation of Harmonic Transfer Functions via Frequency Domain System Identification}
\label{sec:system_id}

%As we noted before, the accuracy and practicality of manually constructed system models are often limited.
In this section, we briefly explain the data-driven system identification method presented in \cite{afreen.msthesis}.

In an LTP system, a sinusoidal input at a specific frequency generates
a superposition of sinusoids at multiple (possibly an infinite number
of) harmonics. Consequently, the system identification framework
starts with truncating the number of harmonic transfer functions
$\htfen$ to be estimated. In the following examples, we consider only
three frequencies in the output to clearly illustrate the approach of
\cite{afreen.msthesis}.

Suppose that an LTP system consists of the superposition of three different
harmonic transfer functions, $\htfen_0$, $\htfen_{-1}$ and
$\htfen_{1}$, each corresponding to a different frequency component of the
output. The output can then be expressed as
\begin{eqnarray}
    \hat{Y}( j \omega ) &=&  \htfen_0( j \omega ) U( j \omega ) + \htfen_{-1}(j \omega) U( j \omega + j \pumpf) \nonumber \\
    & & + \htfen_{1}(j \omega) U( j \omega - j \pumpf).
    \label{eq:output}
\end{eqnarray}

In this new formulation, the $n^{th}$ transfer function is defined as
the linear operator that maps the output at frequency $\omega$ to an
input at the same frequency, modulated with $e^{j n \pumpf
  t}$. However, a single input--output pair in each frequency will
naturally not be sufficient to estimate harmonic transfer functions as
in the case of LTI systems, since the identification problem will then
be underdetermined. Therefore, either at least three independent
inputs or additional constraints must be provided to enable a
successful identification of these harmonic transfer functions.

There are two key issues that need to be addressed before designing
input signals for the identification process. First, we will require
the use of at least as many variations on the input signal as the
number of harmonic transfer functions to be estimated. This is
accomplished in \cite{afreen.msthesis}, which uses a single input
sequence signal for system identification, constructed by
concatenating phase shifted copies of a single waveform on the input
evenly separated by delays within the system period. A complete
characterization of system dynamics is possible with this method since
different modes of the system were activated through the use of
phase-shifted copies of a single waveform.

The second issue is the need to excite all frequency components within
the system by providing input signals with a sufficiently wide
frequency spectrum. This can be accomplished through the use of chirp
signals, whose frequency varies with time. The use of chirp input
signals, combined with the idea of supplying multiple, phase-shifted
input sequences allows us to obtain sufficiently rich input--output data to
support the system identification process.

Using this data with input--output pairs, one can estimate the
harmonic transfer functions of the system, so that the error between
actual and estimated outputs is minimized. Therefore, we can convert
the identification problem to an optimization formulated as
\begin{equation}
\begin{aligned}
        \label{eq:costfunc1}
	& \underset{\htfe}{\text{minimize}}
	& & \mathbf{J} = (\mathbf{Y} - \mathbf{U}^\mathbf{T} \htfe)^2. \\
\end{aligned}
\end{equation}

However, note that Siddiqi \cite{afreen.msthesis} combines all
phase-shifted signals in a single input. Hence, the problem is still
underdetermined in the frequency domain, since a single
input--output pair for a specific frequency will not be sufficient to
identify three harmonic transfer functions. In order to address this
problem, they consider additional constrains on the estimated harmonic
transfer functions. First, they assume that transfer functions are
smooth, which is reasonable for physical systems. This is enforced
through a difference operator, $\mathbf{D^2}$, designed to compute the
second derivative of a vector when multiplied from the left
side. Details on the derivations for $\mathbf{D^2}$ can be found in
\cite{afreen.msthesis}.

The smoothness condition on transfer functions requires penalizing the
curvature of individual transfer functions. Therefore,
\cite{afreen.msthesis} modifies \req{eq:costfunc1} to include a cost
associated with the curvature, yielding a revised minimization problem
formulated as
\begin{equation}
\begin{aligned}
        \label{eq:costfunc2}
        & \underset{\htfe}{\text{minimize}}
	& & \mathbf{J} = (\mathbf{Y} - \mathbf{U}^\mathbf{T} \htfe)^2 + (\alpha \mathbf{D^2} \htfe)^2 \;, \\
\end{aligned}
\end{equation}
\noindent where $\alpha$ is a manually tuned constant weight for
penalizing curvature.  The solution of \req{eq:costfunc2} can easily be
obtained by differentiating $\mathbf{J}$ with respect to
$\htfe$, taking the form
\begin{equation}
\begin{aligned}
        \label{eq:costfuncsol}
        \htfe = (\mathbf{U}^\mathbf{T} \mathbf{U} + \alpha \mathbf{D^4})^{-1} \mathbf{U}^\mathbf{T} \mathbf{Y} \;,
\end{aligned}
\end{equation}
\noindent where the rows of the matrix $\htfe (\omega)$ correspond to
individual different harmonic transfer functions as
\begin{equation}
\htfe (\omega) =
\begin{bmatrix}
\htfe_1 (\omega) \\
\htfe_0 (\omega) \\
\htfe_{-1} (\omega) \\
\end{bmatrix}.
\end{equation}

Note that $\mathbf{U}^\mathbf{T} \mathbf{U}$ and
$\mathbf{U}^\mathbf{T} \mathbf{Y}$ correspond to power spectral and
cross spectral density functions, respectively. Therefore,
\req{eq:costfuncsol} is analogous to estimating transfer functions in
LTI systems, with an additional cost on curvature.

\section{Simplified Legged Locomotion Model with Hybrid System Dynamics}
\label{sec:simplified_model}

In this section, we describe a simple, vertically constrained
spring-mass-damper system that possesses hybrid structural properties
similar to the extensively studied Spring-Loaded Inverted Pendulum
(SLIP) model for running behaviors. This will provide a simple example
to illustrate the application of our system identification method to
such systems.

\subsection{System Dynamics}
\label{sec:system_dynamics}

\begin{figure}[ht]
\begin{center}
\includegraphics[width=0.5\columnwidth]{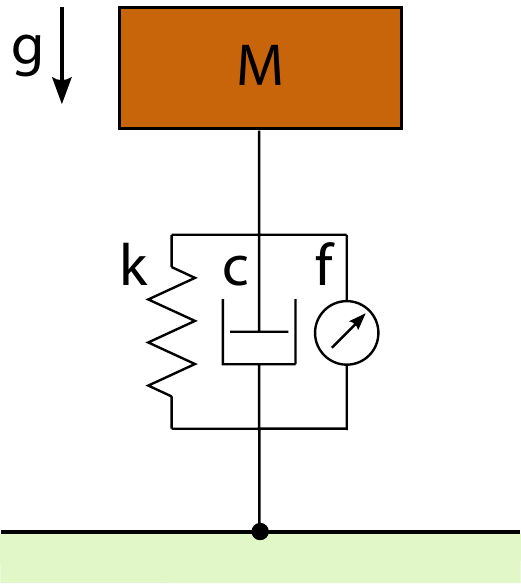}
\caption{Simplified leg model.}
\label{fig:model}
\end{center}
\end{figure}

Fig. \ref{fig:model} illustrates the vertical leg model we focus on in
this section. It consists of a mass attached to a leg with a
spring-damper mechanism as well as a force transducer. Unlike the SLIP
model, we assume that the toe is permanently affixed on the ground.
Nevertheless, we recover the hybrid nature of locomotory gaits by
assuming that the damper is turned on during a ``stance phase''
(lossy) and off during a ``flight phase'' (lossless). This
construction recovers the hybrid nature of the dynamics, while
allowing active input throughout the entire trajectory to support the
generation of system identification data, as well as admitting
theoretical computation of its harmonic transfer functions for a
comparative investigation.

We use the force transducer $f$ in this system for two
purposes. Firstly, active energy input to the system must be provided
to maintain the limit cycle and compensate for energy losses due to
the presence of damping. Second, it will be used as an exogenous input
to the system to support the system identification process. Many
physical legged platforms include similar active components in their
legs to regulate their mechanical energy
\cite{byl.icra2013,gorkem}. Notwithstanding differences in how these
actuators are incorporated into the system, they can all be used as
the necessary exogenous inputs to perform system identification. A
similar model was also investigated in \cite{allen.nonlinear} but
using an additional nonlinear spring for energy regulation.

The equations of motion for this simplified legged locomotion model
are given by
\begin{equation}
\label{eq:eqnofmotion}
     m \ddot{x} =
    \begin{cases}
    -m g -c \dot{x} - k (x - x_0) + \pumping, & \text{if } \dot{x} > 0 \\
    -g - k (x - x_0) + \pumping, & \text{otherwise}.
    \end{cases}
\end{equation}
The lossy and lossless dynamics in \req{eq:eqnofmotion} correspond to
different charts in \refig{fig:statetransition} and zero crossings of
$\dot{x}$ represent threshold functions for both phases.

Our illustrative examples use the parameters $g = 9.81$, $k = 200$, $c
= 2$, $m = 1$ and $x_0 = 0.2$, chosen to be similar to the parameters
of a vertical hopper platform in our laboratory
\cite{uyanik.msthesis}. As noted above, we choose the linear actuator
input $\pumping = \forcing + \chirp$, consisting of a forcing term
$\forcing$ to compensate for energy losses, and a chirp signal
$\chirp$ to introduce small periodic perturbations for system
identification.

\subsection{Theoretical Computation of Harmonic Transfer Functions}
\label{sec:theoretical_htfs}

The goal of this section is to compute harmonic transfer functions
for our model around its limit cycle as outlined in
\resec{sec:htf_theoretical}.

We first assume that the forcing input $\forcing$ is appropriately
chosen to induce an asymptotically stable limit cycle for this
system. For example, our simple leg model achieves a stable limit
cycle with $\forcing = \cos(2 \pi t)$. At this point, changing into
error coordinates away from the limit cycle with $\xi = x(t) -
\bar{x}(t)$, and substituting into \req{eq:eqnofmotion}, the equations
of motion take the form
\begin{equation}
\label{eq:errordynamics}
    \ddot{\xi} =
    \begin{cases}
    -c \dot{\xi} - k \xi , & \text{if } \dot{\xi} + \dot{\bar{x}}(t) > 0 \\
    - k \xi, & \text{otherwise}
    \end{cases}
\end{equation}
Due to the simplicity of the dynamics, this corresponds to a piecewise
LTI system without necessitating any additional approximations, taking
the form
\begin{equation}
    \label{eq:statespace}
    \begin{bmatrix}
    \dot{\xi_1}\\
    \dot{\xi_2}\\
    \end{bmatrix}
    =
    \begin{bmatrix}
    0 & 1 \\
    -k & -c s(\dot{\xi},t)\\
    \end{bmatrix}
    \begin{bmatrix}
    \xi_1\\
    \xi_2\\
    \end{bmatrix}
    +
     \begin{bmatrix}
    0\\
    1\\
    \end{bmatrix}
    \chirp ,
\end{equation}
where the hybrid nature of the system is captured by the flag
$s(\dot{\xi},t)$, with $s = 1$, when $\dot{\xi} + \dot{\bar{x}}(t) >
0$ and $s = 0$ otherwise.

We now need to represent this piecewise LTI system as a linear time
periodic system. However, even though the binary valued function
$s(\dot{\xi},t)$ can be considered time-periodic on the limit cycle
itself, this is not the case for trajectories away from the limit
cycle. To proceed, we hence assume that input induced perturbations
are small, and that the binary valued function $s(\dot{\xi},t)$
maintains its period and becomes strictly time dependent rather than
state dependent, taking the form $s(\dot{\xi},t) \approx s(t)$. We now
can perform a Fourier series expansion on $s(t)$ by treating it as a
square wave with an offset to obtain a linear time periodic system in
the form
\begin{eqnarray}
    \label{eq:statespace2}
    \begin{bmatrix}
    \dot{\xi_1}\\
    \dot{\xi_2}\\
    \end{bmatrix}
    & = &
    \begin{bmatrix}
    0 & 1 \\
    -k & -c s(t)\\
    \end{bmatrix}
    \begin{bmatrix}
    \xi_1\\
    \xi_2\\
    \end{bmatrix}
    +
     \begin{bmatrix}
    0\\
    1\\
    \end{bmatrix}
    \chirp , \\
    \notag
    y  & = &
    \begin{bmatrix}
    1 & 0\\
    \end{bmatrix}
    \begin{bmatrix}
    \xi_1\\
    \xi_2\\
    \end{bmatrix}\;.
\end{eqnarray}
Plugging these equations into the HTF framework described in
\resec{sec:htf_theoretical}, yields analytic solutions to the harmonic
transfer functions. We omit the details of this derivation due to
space considerations, but use the resulting analytic solutions for the
harmonic transfer functions up to $n_h=10$ to evaluate the output of our system
identification method.

\subsection{Data-Driven Identification of Harmonic Transfer Functions}
\label{system_identification}

In this section, we obtain harmonic transfer functions corresponding
to the linearized dynamics of \req{eq:statespace2} by using
input--output data without assuming prior knowledge of the state space
model. Using $\forcing = \cos(2 \pi t)$ and $\chirp = 0$ for 30 cycles
without a perturbation, our example system stabilized to a limit cycle
$\bar{x}(t)$ with a period $T=1s$. We use the $30^{th}$ period as the
numerical limit cycle of the nonlinear system and subtract it from the
trajectories of subsequent experiments to obtain the error function $\xi_1$.

%\njccom{What is ``0Hz'' -- there must be some lowest non-zero
%  frequency, no?} \uscom{We can probably give the formula here, which would make this disappear. Something like: $ u(t) = 0.004 \sin( w(t) t)$, where $w(t) = 14 \pi t/30$? assuming $0\leq t \leq 30$}
In order to obtain input--output data for system identification, we
apply an input signal consisting of nine subsequent $30s$ long chirp
signals, each with a linearly increasing frequency in the range $(0,7]$ Hz
over its duration but with a different starting phase evenly
distributed across the system's period, $T=1s$. Each chirp signal has
an amplitude of $0.004$, chosen to be large enough to perturb system
dynamics but small enough to keep the system close to the periodic
orbit. A sample chirp signal with zero phase can be generated by
\begin{equation}
    u(t) = 0.004 \sin( 14 \pi t^2 / 30).
\end{equation}

\begin{figure}[ht]
\begin{center}
\includegraphics[width=1.0\columnwidth]{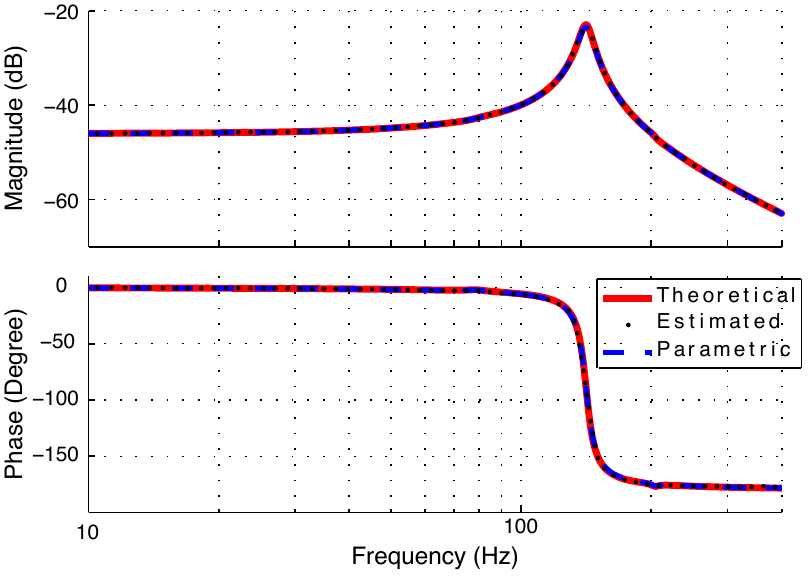}
\caption{Estimation results for the fundamental harmonic.}
%\njccom{This
%  figure's fonts are way too small. Likewise for the next one. Create
%  the page in Adobe Illustrator at the correct size (3.25 inches?), and then the
%  minimum font size should be 7 point, ideally 8.}}
\label{fig:fundamental_harmonic_results}
\end{center}
\end{figure}

The resulting output is then subtracted from the numerically measured
limit cycle to obtain error trajectories $\xi_1$ for vertical
position.  The input signal and $\xi_1$ are then used as in
\resec{sec:system_id} to estimate harmonic transfer functions for our
system. Since our theoretical computations showed that responses beyond
the third harmonic were very small, we only consider the fundamental
harmonic and three harmonics on both sides for our experiments.

\refig{fig:fundamental_harmonic_results} illustrates the estimation
performance of our algorithm for the magnitude and phase of the
fundamental harmonic. Both graphs show that the application of the
identification algorithm in \cite{afreen.msthesis} works well even for
nonlinear periodic systems with hybrid dynamics.

\begin{figure*}[ht]
\begin{center}
\includegraphics[scale=1]{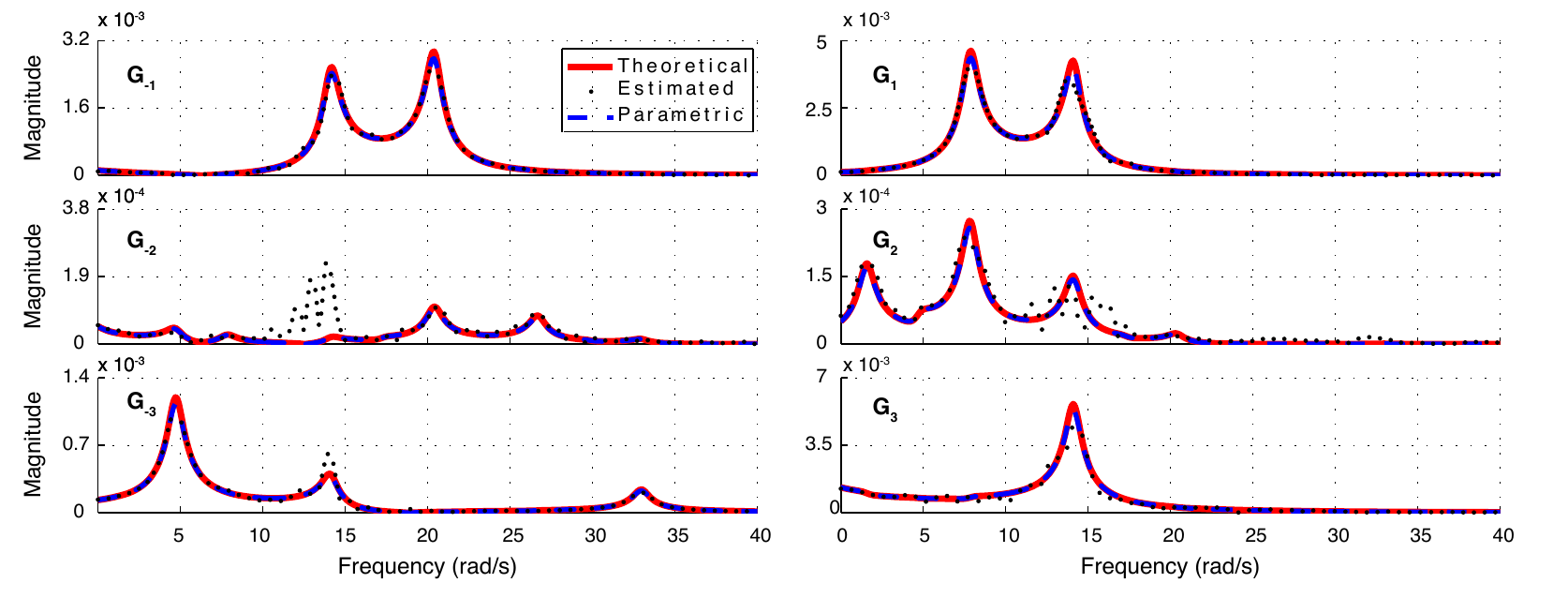}
\caption{Estimation results for the higher order harmonics.}
\label{fig:harmonic_results}
\end{center}
\end{figure*}

We also show our identification results for three harmonics in both
the negative and positive sides in \refig{fig:harmonic_results}. Even
though magnitudes for the harmonic transfer functions are small
compared to the fundamental, the identification algorithm can provide
accurate estimates for these transfer functions except in some narrow
regions of $G_{-2}$ and $G_{2}$. The identification algorithm could
not correctly estimate these two harmonics around $12-15$ (rad/s). One
possible reason for this discrepancy is the presence of strong
responses in all harmonics around the same frequency except $G_{-2}$
and $G_{2}$, resulting in the inability of the identification
algorithm to distinguish between the contributions from each harmonic
absent knowledge of the internal system dynamics. Alternatively, these
discrepancies may also be a result of the fact that hybrid transitions
are not strictly time periodic (rather, they are state-dependent)
which likely has effects on different frequencies and harmonics. We
plan on investigating these issues further in the future.

For a comparative analysis, we also present results from a parametric
identification in order to show that further corrections on estimation
results from a non-parametric method are possible. To this end, we fit
the system parameters $k$ and $c$ in \req{eq:statespace2} by comparing
root mean square error between theoretically computed and estimated
harmonic transfer functions $G_{0}$, $G_{-1}$ and $G_{1}$. We truncate
the system response after the first harmonic in order to discard
erroneous regions in higher harmonics. The resulting estimates were
$\hat{k} = 200$ for the spring constant and $\hat{c}=2.12$ for the
damping coefficient, which closely coincide with the parameters used
to generate the input--output data. As such, harmonic transfer
functions obtained from parametric identification were found to
closely match those obtained from theoretical computations as seen in
\refig{fig:harmonic_results}.

Motivated by these identification results, we plan to extend our work
to the identification of the Spring-Loaded Inverted Pendulum (SLIP)
model \cite{SchwindPhD98} and its extensions, widely used as models of
locomotory behaviors in the literature.  Our future goal is to apply
our system identification methods to our physical monopod robot
platform and to compare the identification performances with our
previously verified analytical model \cite{uyanik.tro}.

\section{Conclusion}

In this paper, we presented a system identification strategy to
estimate input--output transfer functions for a simple hybrid spring
mass damper system as a step towards data-driven models for legged
locomotion. We first showed that a class of hybrid locomotion models
can be approximated with a piecewise constant LTP systems in close
proximity to their asymptotically stable limit-cycle.  Our analysis
and identification framework is based on the concept of harmonic
transfer functions \cite{wereley.1990}.

We first observed that the hybrid system dynamics associated with this
model exhibits piecewise LTI behavior around its periodic orbit. We
then represented this behavior as a purely time periodic system around
the limit cycle in order to utilize system identification techniques
applicable to Linear Time Periodic systems.

In order to provide a basis for comparison, we computed analytic
expressions for harmonic transfer functions associated with the LTP
approximation to our simplified hybrid model. In our theoretical
analysis, we considered the system's response up to the $10^{th}$
harmonic. We observed that there were no meaningful responses on both
positive and negative sides after the third harmonic. Therefore, we
decided to truncate the system response after the third harmonic
during our identification studies.

We then performed systematic simulation studies and identified the
harmonic transfer functions of the same model without knowledge of its
internal dynamics. We used an input signal consisting of successive
chirp signals, with phases evenly distributed across the system's
period, to obtain a full characterization of system dynamics for our
frequency range of interest. Our studies showed that LTP system
identification techniques can successfully be used to identify the
transfer functions of nonlinear periodic models with hybrid system
dynamics.

%%%%%%%%%%%%%%%%%%%%%%%%%%%%%%%%%%%%%%%%%%%%%%%%%%%%%%%%%%%%%%%%%%%%%%%%%%%%%%%%%
\section*{Acknowledgment}

This material is based on work supported by the National Science Foundation (NSF) Grants 0845749 and 1230493 (to N. J. Cowan). The authors thank Aselsan and The Scientific and Technological Research Council of Turkey (T\"{U}B\.{I}TAK) for \.{I}smail Uyan{\i}k's financial support.

%%%%%%%%%%%%%%%%%%%%%%%%%%%%%%%%%%%%%%%%%%%%%%%%%%%%%%%%%%%%%%%%%%%%%%%%%%%%%%%%%

\IEEEtriggeratref{20}
\bibliographystyle{IEEEtran}
\bibliography{limbs-strings,references}

\begin{thebibliography}{10}
\providecommand{\url}[1]{#1}
\csname url@samestyle\endcsname
\providecommand{\newblock}{\relax}
\providecommand{\bibinfo}[2]{#2}
\providecommand{\BIBentrySTDinterwordspacing}{\spaceskip=0pt\relax}
\providecommand{\BIBentryALTinterwordstretchfactor}{4}
\providecommand{\BIBentryALTinterwordspacing}{\spaceskip=\fontdimen2\font plus
\BIBentryALTinterwordstretchfactor\fontdimen3\font minus
  \fontdimen4\font\relax}
\providecommand{\BIBforeignlanguage}[2]{{%
\expandafter\ifx\csname l@#1\endcsname\relax
\typeout{** WARNING: IEEEtran.bst: No hyphenation pattern has been}%
\typeout{** loaded for the language `#1'. Using the pattern for}%
\typeout{** the default language instead.}%
\else
\language=\csname l@#1\endcsname
\fi
#2}}
\providecommand{\BIBdecl}{\relax}
\BIBdecl

\bibitem{holmesdynamics2006}
P.~J. Holmes, R.~J. Full, D.~E. Koditschek, and J.~Guckenheimer, ``The dynamics
  of legged locomotion: Models, analyses, and challenges,'' \emph{SIAM Rev},
  vol.~48, no.~2, pp. 207--304, 2006.

\bibitem{fulltemplates1999}
R.~J. Full and D.~E. Koditschek, ``Templates and anchors: neuromechanical
  hypotheses of legged locomotion on land,'' \emph{J Exp Biol}, vol. 202,
  no.~23, pp. 3325--3332, 1999.

\bibitem{SchwindPhD98}
W.~J. Schwind, ``Spring loaded inverted pendulum running: A plant model,'' {PhD
  Thesis}, University of Michigan, Ann Arbor, MI, USA, 1998.

\bibitem{fullmechanics1991}
R.~J. Full and M.~S. Tu, ``Mechanics of a rapid running insect: two-, four-,
  and six-legged locomotion,'' \emph{J Exp Biol}, vol. 156, pp. 215--231, 1991.

\bibitem{Holmes90}
P.~{Holmes}, ``Poincar{\'e}, celestial mechanics, dynamical-systems theory and
  ``chaos''.'' \emph{Physics Reports}, vol. 193, pp. 137--163, September 1990.

\bibitem{schwind.jnls00}
W.~J. Schwind and D.~E. Koditschek, ``Approximating the stance map of a 2 dof
  monoped runner,'' \emph{Journal of Nonlinear Science}, vol.~10, no.~5, pp.
  533--588, July 2000.

\bibitem{geyer.jtb05}
H.~Geyer, A.~Seyfarth, and R.~Blickhan, ``Spring-mass running: Simple
  approximate solution and application to gait stability,'' \emph{Journal of
  Theoretical Biology}, vol. 232, no.~3, pp. 315--328, February 2005.

\bibitem{ndpaper}
U.~Saranli, O.~Arslan, M.~M. Ankaral{\i}, and {\"{O}}.~Morg\"{u}l,
  ``Approximate analytic solutions to non-symmetric stance trajectories of the
  passive spring-loaded inverted pendulum with damping,'' \emph{Nonlinear
  Dynamics}, vol.~62, pp. 729--742, December 2010.

\bibitem{ankaralistride2010}
M.~M. Ankarali and U.~Saranli, ``Stride-to-stride energy regulation for robust
  self-stability of a torque-actuated dissipative spring-mass hopper,''
  \emph{Chaos: An Interdisciplinary Journal of Nonlinear Science}, vol.~20,
  no.~3, p. 033121, September 2010.

\bibitem{uyanik_saranli_morgul.icra2011}
I.~Uyanik, U.~Saranli, and {\"{O}}.~Morg\"{u}l, ``Adaptive control of a
  spring-mass hopper,'' in \emph{IEEE International Conference on Robotics and
  Automation (ICRA)}, 2011, pp. 2138--2143.

\bibitem{uyanik.tro}
I.~Uyanik, {\"{O}}.~Morg\"{u}l, and U.~Saranli, ``Experimental validation of a
  feed-forward predictor for the spring-loaded inverted pendulum template,''
  \emph{IEEE Transactions on Robotics}, 2015.

\bibitem{saranlirhex:2001}
U.~Saranli, M.~Buehler, and D.~E. Koditschek, ``{RHex}: A simple and highly
  mobile robot,'' \emph{International Journal of Robotics Research}, vol.~20,
  no.~7, pp. 616--631, July 2001.

\bibitem{wereley.phdthesis}
N.~W. Wereley, ``Analysis and control of linear periodically time varying
  systems,'' Ph.D., Massachusetts Institute of Technology, Dept. of Aeronautics
  and Astronautics, 1991.

\bibitem{afreen.msthesis}
A.~Siddiqi, ``Identification of the harmonic transfer functions of a helicopter
  rotor,'' M.Sc., Massachusetts Institute of Technology, Dept. of Aeronautics
  and Astronautics, 2001.

\bibitem{hwang.phdthesis}
S.~Hwang, ``Frequency domain system identification of helicopter rotor dynamics
  incorporating models with time periodic coefficients,'' Ph.D., Graduate
  School of the University of Maryland at College Park, 1997.

\bibitem{allen.nonlinear}
M.~W. Sracic and M.~S. Allen, ``Method for identifying models of nonlinear
  systems using linear time periodic approximations,'' \emph{Mechanical Systems
  and Signal Processing}, vol.~25, no.~7, pp. 2705 -- 2721, 2011.

\bibitem{ankaralisystem2014}
M.~M. Ankarali and N.~J. Cowan, ``System identification of rhythmic hybrid
  dynamical systems via discrete time harmonic transfer functions,'' in
  \emph{Proc IEEE Int Conf on Decision and Control}, Los Angeles, CA, USA,
  December 2014.

\bibitem{gallowayx-rehx:2010}
K.~C. Galloway, G.~C. Haynes, B.~D. Ilhan, A.~M. Johnson, R.~Knopf, G.~Lynch,
  B.~Plotnick, M.~White, and D.~E. Koditschek, ``X-rhex: A highly mobile
  hexapedal robot for sensorimotor tasks,'' University of Pennsylvania, Tech.
  Rep., 2010.

\bibitem{saranlimodular2011}
U.~Saranli, A.~Avci, and M.~C. {\"{O}}zt\"{u}rk, ``A modular real-time fieldbus
  architecture for mobile robotic platforms,'' \emph{IEEE Transactions on
  Instrumentation and Measurement}, vol.~60, no.~3, pp. 916--927, 2011.

\bibitem{revzenestimating2008}
S.~{Revzen} and J.~M. {Guckenheimer}, ``Estimating the phase of synchronized
  oscillators,'' \emph{Physical Review}, vol.~78, no.~5, p. 051907, November
  2008.

\bibitem{guckenheimerplanar1995}
J.~Guckenheimer and S.~Johnson, ``Planar hybrid systems,'' in \emph{Hybrid
  systems II}.\hskip 1em plus 0.5em minus 0.4em\relax Springer, 1995, pp.
  202--225.

\bibitem{wu3_D2013}
A.~Wu and H.~Geyer, ``The 3-d spring--mass model reveals a time-based deadbeat
  control for highly robust running and steering in uncertain environments,''
  \emph{IEEE Transactions on Robotics}, 2013.

\bibitem{leonhard}
A.~Leonhard, ``The describing function method applied for the investigation of
  parametric oscillations,'' in \emph{Proceedings of the 2nd IFAC World
  Conference}, 1963, pp. 21--28.

\bibitem{wereley.1990}
N.~Wereley and S.~Hall, ``Frequency response of linear time periodic systems,''
  in \emph{Proceedings of the 29th IEEE Conference on Decision and Control},
  1990, pp. 3650--3655 vol.6.

\bibitem{mollerstedt.phdthesis}
E.~M\"{o}llerstedt, ``Dynamic analysis of harmonics in electrical systems,''
  Ph.D., Lund Institute of Technology, Department of Automatic Control, 2000.

\bibitem{byl.icra2013}
G.~Piovan and K.~Byl, ``Two-element control for the active slip model,'' in
  \emph{IEEE International Conference on Robotics and Automation (ICRA)}, May
  2013, pp. 5656--5662.

\bibitem{gorkem}
G.~Secer and U.~Saranli, ``Control of monopedal running through tunable
  damping,'' in \emph{Signal Processing and Communications Applications
  Conference (SIU), 2013 21st}, April 2013, pp. 1--4.

\bibitem{uyanik.msthesis}
I.~Uyanik, ``Adaptive control of a one-legged hopping robot through dynamically
  embedded spring-loaded inverted pendulum template,'' {M.Sc.}, Bilkent Univ.,
  Ankara, Turkey, August 2011.

\end{thebibliography}

\end{document}